\documentclass{article} 
\usepackage{iclr2027_conference,times}

\usepackage{url}

\usepackage[utf8]{inputenc} 
\usepackage[T1]{fontenc}    
\usepackage{hyperref}       
\usepackage{url}            
\usepackage{booktabs}       
\usepackage{amsfonts}       
\usepackage{nicefrac}       
\usepackage{microtype}      
\usepackage{xcolor}         

\usepackage{nicefrac}
\usepackage{float}
\usepackage{subcaption}
\usepackage{multirow}
\usepackage{adjustbox}
\usepackage[linesnumbered,ruled,vlined]{algorithm2e}
\usepackage{soul}
\usepackage{marvosym}
\usepackage{amsfonts}
\usepackage{amsmath}
\usepackage{xcolor}
\usepackage{bbm}
\usepackage{siunitx}
\usepackage{lipsum}
\usepackage{tabularx}
\usepackage{algorithmic}
\usepackage{threeparttable}
\usepackage{titletoc}
\usepackage{booktabs}
\usepackage{makecell}
\usepackage{dsfont}
\usepackage{array}

\usepackage{booktabs}
\usepackage{array}
\usepackage{listings}
\usepackage{xcolor}
\usepackage{pifont}
\newcommand{\cmark}{\ding{51}} 
\newcommand{\xmark}{\ding{55}} 

\definecolor{darkgreen}{rgb}{0.0, 0.7, 0.0}  
\definecolor{darkred}{rgb}{0.7, 0.0, 0.0} 
\definecolor{correctgreen}{RGB}{0, 128, 0}
\definecolor{wrongred}{RGB}{180, 0, 0}

\newcolumntype{M}[1]{>{\centering\arraybackslash}m{#1}}

\newtheorem{proposition}{Proposition}

\title{
Alliance of Stochastic Routing and Reinforcement Fusion for Multi-Model Reasoning}

\author{Selim Furkan Tekin\textsuperscript{1}, Ling Liu\textsuperscript{1} \\
\textsuperscript{1} Georgia Institute of Technology}

\iclrfinalcopy 
\begin{document}

\maketitle

\begin{abstract}

This paper studies dynamic model routing under uncertainty in multimodal settings, where a system must select and potentially compose multiple models to answer a query. Standard reinforcement learning approaches for routing often suffer from premature convergence to suboptimal models due to noisy rewards and insufficient exploration. We propose a unified vision-language {\bf S}to{\bf c}hastic r{\bf out}ing framework (VL-Scout) for efficient multi-Modal and multi-step reasoning with turn-based reinforcement fusion with three unique features. First, 
VL-Scout combines Bayesian priors, policy gradient optimization, and sequential decision-making to enable structured exploration and adaptive inference. A context-dependent Dirichlet belief model allows for epistemic uncertainty over model quality and is updated online based on observed rewards and context-specific evidence, enabling Thompson-style exploration while serving as a prior for regularization. Second, a routing policy is trained via a KL-regularized REINFORCE objective by maximizing a reward-weighted evidence lower bound, where the policy acts as an amortized posterior, and the Dirichlet belief serves as a prior. Third, we introduce a fusion policy trained with Group Relative Policy Optimization (GRPO) or Reinforcement Learning with Verifiable Rewards (RLVR), which operates over intermediate model rationales and selects between two actions: CONTINUE, invoking another model via the router, or ANSWER, terminating with a generated output. This enables adaptive computation depth and selective model composition conditioned on reasoning quality. We evaluate VL-Scout on four representative multi-modal benchmarks (MMMU, MMMU-Pro, OKVQA, and RealWorldQA), compared against 11 representative routing and policy optimization baselines. Our method provides consistent performance improvements, while reducing premature policy collapse. These results highlight the importance of integrating uncertainty-aware priors with both amortized policy learning and sequential fusion for robust and efficient model selection in discrete action spaces. Code is \href{https://github.com/sftekin/scout-rl/tree/release}{available}

\end{abstract}

\vspace{-4pt}
\section{Introduction}
\label{rl_focal:Introduction}
\vspace{-6pt}

Multi-modal reasoning is recognized as a challenging problem, as it requires models to understand, interpret, and process information from diverse sources. Advanced visual-language models (VLMs) have shown remarkable performance across different tasks, domains, and query types. However, no single model to date  dominates uniformly across all evaluation settings and benchmarks. 

Existing research tackles such fundamental heterogeneity in model competencies along two orthogonal directions:
Model routing \citep{ong2024routellm, zhang2025router, li2025generation} aims to select the right model for each query, aiming for low cost and high accuracy. However, model routing fails to fully utilize the diversity in model capabilities, as shown by the recent progress in ensemble learning of LLMs/VLMs  \citep{tekin2025dynamic, wan2024knowledge, yu2024breaking, huang2024ensemble, mavromatis2024pack, yao2024determine, jiang2023llm, tekin-etal-2024-llm} and collaboration \citep{sivakumaran2025dart, liang2024encouraging, du2023improvingfactualityreasoninglanguage}. 
Little efforts have targeted at tackling both routing and fusion  together. Recently, \citep{nguyen2024metallm} modeled routing as a multi-armed bandit with contextual information and employed a linear relation. However, visual-language modality is too complex to learn through a linear relation. \citep{wang2025mixllm} shows that a dynamic routing with a model pool can be maintained through continual learning. \citep{tekin2025dynamic} utilized reinforcement learning to select LLMs from a model pool and ensemble their logits. However, the introduced algorithm requires a warm-start dataset and a common value network to stabilize the agents (see Appendix for more detail).

In this paper, we propose {\sc VL-Scout}, an online routing and fusion framework for vision-language models. It aims to combine model routing and model fusion by jointly optimizing both routing function and fusion function together in a dynamic changing environment where supervision is rarely available. We advocate that a practical routing system must be capable of learning effective routing policies online and adapt to new query distributions without relying on pre-collected task-specific training data. Leveraging this routing capability, an efficient ensemble fusion system should learn how to fuse conflicting outputs from multiple diverse base learners, regardless of which of the ensemble sets is chosen to solve a reasoning problem.

The design of VL-Scout has three salient features. 
First, we formulate routing as a multi-armed bandit problem and learn a non-linear mapping between model reward distributions and encoded visual-language context via a neural network, which is regularized by a Thompson-sampled prior from a category-conditioned Dirichlet belief on the same simplex and serves as global memory over the history of model performance across task categories. Second, we leverage train a routing policy via a KL-regularized REINFORCE objective by maximizing a reward-weighted evidence lower bound. Third, we propose a fusion model that operates on the routed model's rationales and outputs to adaptively decide whether to consult additional models or generate a final answer directly, enabling selective model composition conditioned on reasoning quality. Extensive evaluations across four benchmark datasets demonstrate that VL-Scout surpasses the best-performing base model, outperforms 11 representative VLMs by up to 5.36\%, and exceeds the generalization performance of five recent ensemble and collaboration approaches by up to 3\% at 8× lower inference cost.

\section{Prior Work}
\label{sec:related_work}


\textbf{Ensemble Learning in Large Vision and Language Models} Ensemble training methods and the distillation
~\citep{wan2024knowledge, yu2024breaking, huang2024ensemble, mavromatis2024pack, yao2024determine} and the mixture of experts (MoE)~\citep{jiang2024mixtral} methods require significant computational resources and full access to the model parameters, making it challenging to generalize to diverse contexts and adapt to domain shifts. In the post-inference aggregation ensemble learning category, most LLM ensemble research centered on supervised solutions~\citep{jiang2023llm, tekin-etal-2024-llm} and utilized majority voting to perform inference-time ensemble~\citep{wang2022self, fu2022complexity, li2022advance, wang2022rationale}. The downside of majority voting is the poor definition of equality between divergent answers. Two threads of research further improve majority voting, one utilizes the BLEU score as the heuristic to compare answers~\citep{li2024more}, and the other enhances the BLEU score-based answer-combination method by assigning weights~\citep{yao2024tree} or creating a debate environment~\citep{liang2023encouraging, wan2024knowledge, du2023improving, chan2023chateval}. LLM-based Multi-Agents \citep{guo2024large} carry similar motivations in terms of exploiting multiple LLMs to work collaboratively for a particular task. However, neither the dynamic selection of agents nor the aggregation of outputs for a given task have been systematically explored.  Similarly. LLM routing aims to identify the most suitable model among the pre-defined set of LLMs for a given prompt query. Unfortunately, routing is inherently limited by the performance of the chosen model and the dependency of using another external LLM to understand/rank the matching of a prompt query to the given pool of LLMs~\citep{ong2024routellm}.  

\textbf{Reinforcement Learning in Model Routing and Ensemble} Recently, RL approaches~\citep{chakraborty2025collab, fu2025rlae} are proposed, which can dynamic adjust ensemble weights for an ensemble of size $N$ ($N$ is fixed for all tasks). Ensemble by RL holds the potential of creating an adaptive ensemble model~\citep{song2023ensemble, chua2018deep}, ranging from time-series prediction~\citep{liu2020new, nemeth2022split, perepu2020reinforcement}, ensemble pruning~\citep{partalas2009pruning, liu2020instance}, to Tree-of-Thought (ToT) family of in-context learning of LLMs~\citep{ouyang2022training, liu2024rl, monea2024llms, zhang2021multi, sun2024llm, liu2024dynamic}. Yet, these existing RL approaches struggle to tackle the challenges when the ensemble of LLMs (base learners) offers very different inference performance with respect to diverse downstream reasoning tasks, given heterogeneous neural architectures finetuned with different LLM serving objectives as well as the challenges of whether it is feasible to dynamically compose an ensemble from a pool of base learners on demand to better address the inference performance demand of each downstream user task.

\vspace{-6pt}
\section{Preliminaries and Motivation}
\label{sec:prem}
\vspace{-6pt}
\textbf{Routing Function.\/} The routing is the subset problem of finding the best possible ensemble set combination for a given query. 
It selects a single model among candidate models. The objective is to learn a function $f_{\mathrm{route}}: \mathcal{X}\rightarrow \{1,\dots,K\}$ that maps a query $x$ to the most suitable foundation model, e.g., LLM, VLM, in a model pool of size $K$ to achieve the best response with the lowest cost. Following the definition \citep{nguyen2024metallm}, the routing functions maximize the performance on $N$ queries with the given budget constraint $b$: 
\begin{equation}
   \arg \max_{f_{\mathrm{route}}}\sum_{i=1}^{N} a_{f_{\mathrm{route}}(x_i)}(x_i)\; \mathrm{s.t.}\; \sum_{i=1}^{N}c_{f_{\mathrm{route}}(x_i)}\leq b .
\end{equation}
More precisely,
by defining $\mathbf{S}\in\{0,1\}^{N\times K}$ and $\mathbf{A}\in\mathbb{R}^{N\times K}$, where $\mathbf{S}_{i,j} = \mathds{1}_{\{f_{\mathrm{route}}(x_i)=j\}}$ is representing the choice of routing function $f_{\mathrm{route}}$ on $N$ queries and $\mathbf{A}_{i,j}=a_j({x_i})$ is the performance of model $j$ on sample $x_i$. The optimization goal is to find a choice matrix that maximizes the overall performance: 
\begin{equation}
   \arg \max_{\mathbf{S}}\sum_{i=1}^{N}\sum_{j=1}^{K} \mathbf{A}_{i,j}\mathbf{S}_{i,j}\; \mathrm{s.t.}\; \sum_{i=1}^{N}c_{f_{\mathrm{route}}(x_i)}\leq b .
\end{equation}
The optimal $\mathbf{S}$ can be reached when the performance (e.g. accuracy) matrix is known \citep{chen2022efficient, chen2023frugalgpt}. Yet, we do not know the accuracy of a model on a test sample before feeding the sample to the model. Even if we know, \textit{the performance feedback may be delayed, and a model’s true performance may only be revealed after multiple queries.} Previous works on LLM-Routing try to overcome the problem by learning a router-model which predicts the performance of each model in the pool. However, this requires the router to comprehend the intricate details of the query and each model's capabilities. The vision modality exacerbates the complexity of untangling these relations. Jointly optimizing these two aspects may limit exploration and adaptive inference, resulting in premature convergence to suboptimal models. Therefore, router models generally require huge parameter sizes or high computational cost due to their cascading query protocols. 
In this work, we formalize both the routing and ensemble problems within a reinforcement learning framework, enabling the learning of more generalizable, \textit{meta-level} strategies.

{\bf Problem Formulation.\/} 
Let $(x, \mathbf{I})$ denote an input tuple of a language query and an image, where the query $x=\{\omega_1, \dots, \omega_L\}$ is composed of $L$ tokens, and the image $\mathbf{I}\in\mathbb{R}^{H\times W\times 3}$ represents the RGB input. Let $(x_1, \mathbf{I}_1), \dots, (x_n, \mathbf{I}_n)$ denote a sequence of input queries with length $n$, where each query $(x_i, \mathbf{I}_i)$ is sampled from a task $\mathcal{T}_j$, denoted by $(x_i, \mathbf{I}) \sim \mathcal{T}_j$.  The queries can originate from a single task or from a group of tasks $\mathcal{T}_1, \mathcal{T}_2, \dots, \mathcal{T}_J$, e.g., spatial-visual reasoning, science-math, and social-applied with varying difficulties. Let $\mathcal{E}_t=\left\{\mathcal{M}_1, \dots, \mathcal{M}_N\right\}$ denote a set of VLMs that can be used to obtain the desired output $y$ at time $t$. For each query $x_i$, the routing aims to minimize $\mathcal{L}\left(\mathcal{M}_{f_{\mathrm{route}}(x_i)}(x), y\right)$ where $\mathcal{M}_{f_{\mathrm{route}}(x_i)}$ is the routed model and $\mathcal{L}$ a generic loss function such as L2. 
The ensemble fusion function objectifies $\mathcal{L}\left(\hat{y}_{\mathrm{fusion}}, y\right)$, where $\hat{y}_{\mathrm{fusion}} = f_{\mathrm{ens}}(\mathcal{M}_1(x_i), \dots, \mathcal{M}_K(x_i))$. 
The design of VL-Scout aims to route and fuse jointly together. On one hand, we want to select the best models for each incoming query with both textual and visual (image) modality; and on the other hand, we want to fuse the selected model outputs with query-aware adaptability. Therefore, for each query, the common objective is formulated as follows: 
\begin{equation}
\min_{\theta,\phi}\mathcal{L}\left(y, f_{\mathrm{ens}}\left(\{\mathcal{M}_k(x_i):k\sim \hat{f}_{\mathrm{route}}(x_i;\theta)\}, \phi\right)\right)\; \mathrm{s.t}\; \sum_{i=1}^{N}c_{f_{\mathrm{route}}(x_i)}\leq b .    
\label{eq:main_objective}
\end{equation}
Here, $\hat{f}_{\mathrm{route}}:\mathcal{X}\rightarrow[0,1]^{K}$ is the routing function that maps the input query to a probability distribution over $K$ models. In other words, here we sample a possible ensemble set from a universal set $\mathbb{U}$ containing all the possible ensemble combinations with $|\mathbb{U}|=2^{K}-K-1$. Our objective is to minimize the loss by finding the parameters $\theta$ and $\phi$ that parametrize the routing and fusion function. Moreover, the optimization has a budget constraint and the feedback signal $y$ can be delayed or noisy. With these elements, the optimization problem distinguishes itself from the literature by its difficulty, modality, and applicability. 

\vspace{-6pt}
\section{VL-Scout}
\vspace{-6pt}

\begin{figure*}[t]
    \centering
    \vspace{-6pt}
    \includegraphics[width=\textwidth]{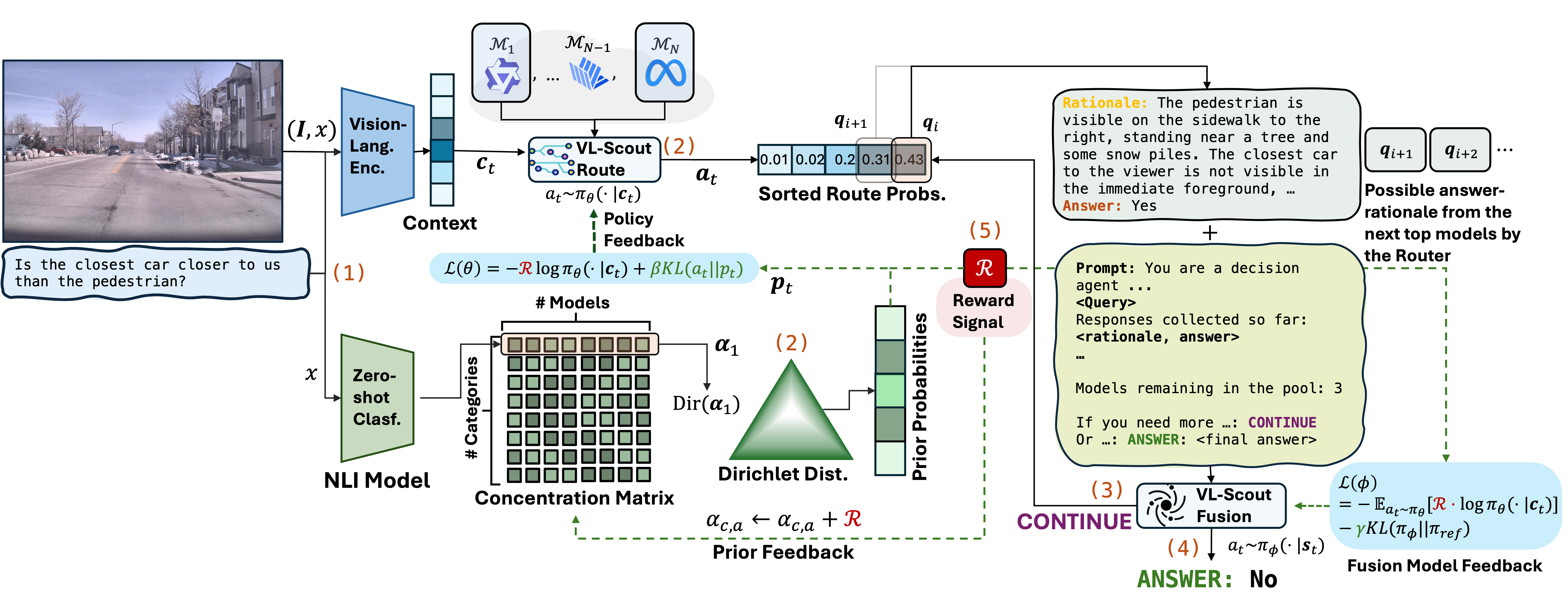}
    \vspace{-5pt}
    \caption{{ Overview of \textit{VL-Scout} two stage ensemble by reinforcement learning agents. The step sequence number is used to annotate the order of the workflow steps of VL-Scout. 
    }}
    \label{fig:main}
    \vspace{-6pt}
\end{figure*}

We first give an architectural overview of VL-Scout in Figure \ref{fig:main}, with a highlight of five steps: (1) The input image and query are passed to the Vision-Language Encoder to obtain its contextual embedding and Natural Language Inference model for zero-shot classification of one of the categories. (2) We obtain probabilities for the most suitable models in the pool based on the context from VL-Scout Route Agent and Prior Distribution. (3) The top selected model's response and rationale is prompted to VL-Scout Fusion Model to generate either two actions CONTINUE, invoking another model via the router, or (4) ANSWER, terminating by generating a final answer. At the final step (5), the reward is calculated based on the final answer and is fed back to the policies and posterior update for new prior. This process iterates until all episodes are finished. In Appendix-\ref{sec:design_guidelines}, we introduce four routing techniques that serve as the foundational components of our proposed algorithm, VL-Scout Route. Presenting these methods provides the necessary context for understanding the design and behavior of our approach.

\subsection{VL-Scout Route: REINFORCE with Prior Regularization}

The Multi-Armed Bandit routers (see Appendix~\ref{sec:design_guidelines}) each solve only part of the problem: Thompson Sampling ignores the query context, LinUCB and its Thompson variant rely on a linear reward model that breaks down under domain shift, and REINFORCE captures non-linear context but cannot use prior knowledge of which models suit which query categories and can converge prematurely to suboptimal models under noisy or delayed rewards, so an effective router must be context-aware, non-linear, prior-guided, and exploratory at the same time. 

\textbf{Context and category.} We assume a predefined set of categories $\mathcal{C}$ to which a query $x \in \mathcal{X}$ may belong (see the list of categories in Appendix-\ref{sec:classification}). For each category $c \in \mathcal{C}$, we maintain a prior belief over the performance of $K$ models, represented as a probability distribution over arms, $p(a|c(x))$. We map each input query to a category using a function $c: \mathcal{X} \rightarrow \mathcal{C}$, instantiated via a Natural Language Inference (NLI) model that assigns the query to the most relevant category with zero-shot classification (see step-1 in Figure \ref{fig:main}). Additionally, the router observes $\mathbf{h}_t = g(x_t, \mathbf{I}_t)\in \mathbb{R}^{768}$, the mean-pooled Q-Former output of frozen BLIP-2 encoder \citep{li2023blip}.

\textbf{Policy.} The router is a categorical policy over the $K$ models, $\pi_\theta(k\mid\mathbf{h}_t)=\mathrm{softmax}(f_\theta(\mathbf{h}_t))_k$, where $f_\theta$ is a two-layer MLP. A top-1 routing decision samples $k_t\sim\pi_\theta(\cdot\mid\mathbf{h}_t)$; a shortlist of length $n$ appends the $n-1$ remaining models in decreasing order of $\pi_\theta$.

\textbf{Belief.} For each category $c$, VL-Scout maintains a Dirichlet belief $\mathrm{Dir}(\boldsymbol{\alpha}_c)$, $\boldsymbol{\alpha}_c\in\mathbb{R}^{K}_{>0}$, over which model to prefer for that category and $\tilde{\mathbf{p}}_t\sim\mathrm{Dir(\boldsymbol{\alpha}_c)}$ is the sampled belief. The entries of $\boldsymbol{\alpha}_c$ act as pseudo-counts: the expected preference for model $k$ is $\alpha_{c,k}/\sum_j\alpha_{c,j}$, and the total $\sum_j\alpha_{c,j}$ sets how firmly the belief is held---the larger it is, the closer the samples drawn from the belief stay to this mean. Giving every model the same initial count $\alpha_0$ yields an uninformative belief. An operator who expects some models to be strong, e.g., from public leaderboard results, can designate them as a set $\mathcal{P}$ and give them a larger initial count $\alpha_{\mathcal{P}}>\alpha_0$; this is how prior knowledge about the pool enters the system. The prior is an adaptive head start of pseudo-counts: after each query, every consulted model $k$ contributes signed evidence
\begin{equation}
\alpha_{c_t,k}\leftarrow\alpha_{c_t,k}+e_{t,k},\qquad e_{t,k}=\begin{cases}1 & \text{if } \hat{y}_{t,k}=y_t,\\ -\lambda_f & \text{otherwise},\end{cases}
\label{eq:belief}
\end{equation}
and concentrations that become non-positive are reset to $\alpha_{\min}>0$. For instance, a non-designated model that answers ten queries of a category correctly reaches $\alpha_{c,k}=11$ and overtakes a designated model that has gained no evidence in that category. Without the failure term ($\lambda_f=0$), Eq.~(\ref{eq:belief}) is the conjugate update of a categorical observation model whose observations are the identities of models that answered correctly; the failure term additionally removes mass from models that are consulted and wrong. The belief is therefore a pseudo-count \emph{preference} over the models of a category, not an exact posterior over each model's success rate. Because the counts are not normalized by how often a model is consulted, the belief favors models that are both often consulted and often correct; exploration comes from sampling it.


\textbf{Objective.} Our goal is to learn a query-dependent routing policy that selects models with high expected reward while remaining robust to noisy and sparse feedback. In particular, we want to avoid two undesirable behaviors: (i) prematurely collapsing onto a model that happens to perform well on a few early queries, and (ii) failing to explore models whose quality is still uncertain. We address these issues by combining a Bayesian belief over model quality with KL-regularized policy optimization: 
\begin{equation}
    J_t(\theta)=\mathbb{E}_{k\sim\pi_{\theta}(\cdot\mid\mathbf{h}_t)}\left[r_t(k)\right] - \beta\mathcal{D}_{KL}(\pi_{\theta}(\cdot\mid\mathbf{h}_t)||\tilde{\boldsymbol{p}}_t),
    \label{eq:loss_route}
\end{equation}
where $r_t(k)=1$ if the answer obtained from the routing decision is correct and $r_t(k)=-0.1$ otherwise. Both $\pi_\theta(\cdot\mid\mathbf{h}_t)$ and $\tilde{\mathbf{p}}_t$ are categorical distributions on the simplex $\Delta^{K-1}$, so the KL term is well defined and computed exactly. Equation~(\ref{eq:loss_route}) is the Lagrangian relaxation of maximizing the expected reward subject to $\mathrm{KL}(\pi_\theta\|\tilde{\mathbf{p}}_t)\le\epsilon$~\citep{boyd2004convex}, with a fixed multiplier $\beta$. Thus, the router seeks high reward while remaining within a controlled distance of the sampled belief. In the following preposition, we show that the posterior is reward-weighted version of the belief by following standard identity~\citep{levine2018reinforcement} which makes precise in what sense the trained router is an amortized posterior.

\begin{proposition}[Gibbs posterior]
\label{prop:gibbs}
Let $\tilde{\mathbf{p}}$ lie in the interior of $\Delta^{K-1}$, $r:\{1,\dots,K\}\to\mathbb{R}$, $\beta>0$, and $\pi^\star(k)=\tilde{p}_k\,e^{r(k)/\beta}/Z$, where the normalizing constant $Z=\sum_{j}\tilde{p}_j\,e^{r(j)/\beta}$ ensures $\sum_k\pi^\star(k)=1$. For every $\pi\in\Delta^{K-1}$,
\begin{equation*}
\mathbb{E}_{k\sim\pi}[r(k)]-\beta\,\mathrm{KL}(\pi\|\tilde{\mathbf{p}})=\beta\log Z-\beta\,\mathrm{KL}(\pi\|\pi^\star),
\end{equation*}
so $\pi^\star$ is the unique maximizer. Moreover, for a binary success variable $O$ with likelihood $P(O{=}1\mid k)\propto e^{r(k)/\beta}$ and prior $\tilde{\mathbf{p}}$ over models, the objective divided by $\beta$ equals, up to an additive constant, the evidence lower bound on $\log P(O{=}1)$ with variational distribution $\pi$, and $\pi^\star$ is the exact posterior $P(k\mid O{=}1)$.
\end{proposition}

The proof of the preposition is in Appendix-\ref{sec:proof}. The trained policy maps $\mathbf{h}_t$ directly to the distribution and amortizes the per-query posterior $\pi_t^*\propto\tilde{\mathbf{p}}_t\,e^{r_t/\beta}$ across queries. The belief proposes, the reward reweights, and the policy generalizes the result through context embedding. Since calculating posterior requires observation of $r_t$ for all the models, we estimate the gradient of the first term in \ref{eq:loss_route} with REINFORCE~\citep{williams1992simple} and differentiate the KL term exactly: 
\begin{equation}
\nabla_\theta J_t=\mathbb{E}_{k\sim\pi_\theta}\big[r_t(k)\,\nabla_\theta\log\pi_\theta(k\mid\mathbf{h}_t)\big]-\beta\sum_{k=1}^{K}\nabla_\theta\pi_\theta(k\mid\mathbf{h}_t)\,\log\frac{\pi_\theta(k\mid\mathbf{h}_t)}{\tilde{p}_{t,k}} .
\label{eq:grad_route}
\end{equation}



\subsection{VL-Scout Fusion}

Routing struggles with the early stopping by picking one model and stopping. The strength of combining multiple model outputs and their relations is never discovered. The routed model may make an error that is irreversible. It is possible that consulting another model yields a correct solution or a different perspective by leveraging complementary strengths of multiple models and their rationales. Use of multiple models, however, creates two additional questions: (i) how to create an ensemble set and (ii) how to combine the models' output with a generalizable solution described in Section \ref{sec:prem}. VL-Scout Fusion introduces a fusion model which is parameterized by a pre-trained VLM that learns to defer decision and refine predictions by outputting CONTINUE, defer decision and gather more evidence, and ANSWER, stop and generate new refined predictions. Due to the autoregressive nature, the VL-Scout can fuse the answers given to the open-ended questions.
As shown in Figure~\ref{fig:main}, the fusion model exploits the rationales of the candidate models, sorted by their routing probabilities. Given the prompt, the fusion model follows a sequential decision process to determine when to stop or ask for more information. The fusion model thus turns static routing into adaptive, multi-step decision-making with learned stopping. The objective can be formalized as an MDP with the following elements. \textbf{State}: $s_t=(x, h_t)$, where $x$ is the input query with its image and $h_t$ is the history up to step $t$, consisting of the selected models, their outputs, and rationales. \textbf{Action}: $a_t\in\{\text{CONTINUE}, \text{ANSWER}\}$. \textbf{Transition}:
\begin{equation}
    s_{t+1} =
    \begin{cases}
        \big(x,\; h_t\cup \mathcal{M}_{j}(x)\big) & \text{if } a_t = \text{CONTINUE}, \\
        \hat{y}_{\mathrm{fusion}} & \text{if } a_t=\text{ANSWER},
    \end{cases}
    \label{eq:fusion_transition}
\end{equation}
where $\mathcal{M}_j$ is the next routed model. \textbf{Reward}: the reward can be binary feedback testing whether $\hat{y}_{\mathrm{fusion}}=y$, or it can be measured by a reward model. To incentivize consulting more models and to respect the budget constraint in Eq.~\eqref{eq:main_objective}, we add the following penalties:
\begin{equation}
    r_t =
    \begin{cases}
        -\dfrac{n}{K}\,\kappa & \text{if } a_t = \text{ANSWER} \text{ and } \hat{y}_{\mathrm{fusion}} \neq y, \\[6pt]
        r' & \text{if } n > n',
    \end{cases}
    \label{eq:fusion_penalty}
\end{equation}
where $n$ is the number of models queried, $\kappa\in[0,1]$ is the early-stop penalty, and $r'$ is a large negative reward for exceeding the budget of $n'$ models; we set $n'=3$ and $r'=-5$. We use model count for coarse budget control; token-level costs enable finer-grained rewards. The \textbf{objective} is to maximise $\mathbb{E}_{\tau\sim \pi_\phi}\left[\mathcal{R}(\tau)\right]$ with trajectory $\tau=(s_1, a_1, \dots, s_T, a_T)$. We adapt two post-training techniques for VL-Scout Fusion.

\textbf{Reinforcement Learning with Verifiable Rewards (RLVR)}~\citep{wen2025reinforcement} defines reward via a verifier function: $\mathcal{R}(\tau) = V(\hat{y}_{\mathrm{fusion}},y)$, where $V\in[0,1]$ is the correctness/accuracy function. Then, the model is trained via policy gradient $\nabla_{\phi}J = \mathbb{E}\left[\mathcal{R}(a,x)\nabla_{\phi}\log\pi_\phi \right]$. RLVR offers a flexible, modular, and simple reward design and directly optimizes the final task objective. However, sparse rewards such as binary may cause a high-variance effecting early actions to get weak signal. 

\textbf{Group Relative Policy Optimization (GRPO)}~\citep{shao2024deepseekmath} reduces the variance by using relative rewards within a group of trajectories. For each query $x$ the policy samples $G$ trajectories $\tau_1,\dots,\tau_G\sim\pi_\phi$ and calculate their rewards. Then the model is trained via GRPO loss function:
\vspace{-3pt}
\begin{equation}
\small
\begin{split}
        \mathcal{L}_{\mathrm{GRPO}}(\phi) = \frac{1}{G}\sum_{i=1}^{G}\frac{1}{|\tau_i|}\sum_{t=1}^{|\tau_i|}&\Big\{\min \Big(
        \hat{r}_{t}(\phi)\hat{A}_{i,t}, 
    \mathrm{clip}\big(\hat{r}_{t}(\phi), 1-\epsilon, 1+\epsilon\big)\hat{A}_{i,t}\Big)-\beta\mathcal{D}_{\text{KL}}(\pi_\phi||\pi_{\text{ref}})\Big\},\\
\end{split}
\end{equation}
where $\hat{r}_{t}(\theta) =\frac{\pi_{\theta_{\mathrm{new}}}(a_t|\mathbf{s}_t)}{\pi_{\theta_{\mathrm{old}}}(a_t|\mathbf{s}_t)}$ is the ratio term, $\hat{A}_t = \mathcal{R}_t-\bar{R}$ is the estimated advantage function to reduce variance and stabilize learning under sparse rewards. GRPO with the ratio term and reference model penalty encourages structured exploration, best when the reward function is relative and noisy. 

\section{Experimental Evaluations}

\begin{table*}[hbt!]
  \vspace{-6pt}
    \captionof{table}{{VL-Scout selection and ensemble composition for the evaluated VLM benchmarks.}}
    \vspace{-5pt}
  \begin{adjustbox}{width=\textwidth, center}
    \centering
    \small
    \begin{tabular}{l l c c c c}
        \toprule
        \multirow{2}{1.5cm}{\textbf{Settings}} & \multirow{2}{*}{\textbf{VLMS}} & \textbf{MMMU} & \textbf{MMMU-Pro} & \textbf{A-OKVQA} & \textbf{RealWorldQA} \\
        \multirow{9}{1.5cm}{Pretrained} & & (Acc \%)$\uparrow$ & (Acc \%)$\uparrow$ & (Acc \%)$\uparrow$ & (Acc \%)$\uparrow$  \\
        \hline
        & llava-v1.6-vicuna-7b & 36.27 & 35.42 & 82.62 & 38.30 \\
        & llava-v1.6-vicuna-13b & 36.77 & 33.06 & 83.41 & 48.37 \\
        & Llama-3.2-11B-Vision & 39.00 & 38.94 & 81.48 & 33.86 \\
        & deepseek-vl2-tiny & 39.75 & 37.96 & 80.09 & 64.2$^*$ \\
        & deepseek-vl2-small & 39.01 & 34.63 & 76.68 & 65.4$^*$ \\
        & Qwen2.5-VL-7B & 51.55 & 46.95 & 87.25 & 69.15 \\
        & Qwen3-VL-8B & 56.64 & 50.56 & 87.94 & 71.11 \\
        & Kimi-K2.5 & 30.18 & 24.87 & 80.96 & 64.97 \\
        & gemma-3n-E4B & 48.57 & 43.03 & 83.06 & 56.47 \\
        & InternVL2-8B & 51.30 & 43.12 & 85.33 & 70.10$^*$ \\
        & InternVL3-8B & 57.76 & 51.82 & 88.47 & 63.53 \\
        \hline
        & \multicolumn{2}{l}{\textbf{\textit{Routing Only (Top-1)}}}  \\
        \multirow{10}{1.5cm}{Online w/o Training Data} & TS-MA & 55.40 & 48.36 & 86.38 & 62.75 \\
        & LinUCB & 48.08 & 42.39 & 84.62 & 59.73\\
        & TS+LinUCB & 49.32 & 44.78 & 88.21 & 57.51\\
        & REINFORCE & 52.91 & 48.05 & 85.15 & 67.45\\
        & VL-Scout Route (REINFORCE+Prior) & 56.65 & 50.13 & 86.72 & 68.15\\
        \cline{2-6}
        & \multicolumn{2}{l}{\textbf{\textit{Routing + Fusion}}}  \\
        & Voting+Bernoulli & 54.04 & 50.34 & 85.76 & 67.45\\
        & Voting-Top-3 & 57.64 & 51.38 & 89.78 & 71.24\\
        & VL-Scout Fusion (RLVR) & 52.67 & 46.76 & 89.00 & 72.55\\
        & VL-Scout Fusion (GRPO) & \textbf{60.37} & \textbf{52.01} & \textbf{90.13} & \textbf{76.47}\\
        \hline
        \multirow{10}{1.5cm}{Online w/ Cross-Domain Training Data} & \textbf{\textit{Routing Only(Top-1)}} & \thead{MMMU-pro\\$\rightarrow$ MMMU} & \thead{MMMU\\$\rightarrow$MMMU-pro} & \thead{RealWorldQA\\$\rightarrow$OKVQA} & \thead{A-OKVQA\\$\rightarrow$RealWorldQA} \\
        \cline{2-6}
        & TS-MA &  55.90 & 50.00 & 87.77  & 69.02 \\ 
        & LinUCB & 41.20 & 41.27 & 74.41 & 56.99 \\ 
        & TS+LinUCB & 35.62 & 36.80 &  72.31 & 58.82 \\ 
        & REINFORCE &  48.57 & 50.65 & 86.37 & 63.52 \\ 
        & VL-Scout Route (REINFORCE+Prior) &  55.78 & 49.82 & 86.20 & 71.11 \\
        \cline{2-6}
        & \multicolumn{2}{l}{\textbf{\textit{Routing + Fusion}}}  \\
        & Voting+Bernoulli  & 54.65 &  51.63   & 87.42   & 68.11 \\
        & Voting-Top-3 & 57.28 &  50.31   & 88.04   & 72.42 \\
        & VL-Scout Fusion (RLVR) & 56.65 &  49.37   & 89.00   & 72.29 \\
        & VL-Scout Fusion (GRPO) & \textbf{62.86} &  \textbf{53.52}   & \textbf{90.57}   & \textbf{74.51} \\
        \hline
        & \textit{Oracle} & 81.98 & 78.14 & 95.89 & 92.54 \\
        \bottomrule    
    \end{tabular}
    \end{adjustbox}
    \label{tab:main}
    \vspace{-4pt}
\end{table*}

We evaluate VL-Scout across four benchmarks in Table \ref{tab:main} — MMMU~\citep{yue2024mmmu}, MMMU-Pro~\citep{yue2024mmmu_pro}, A-OKVQA~\citep{schwenk2022okvqa}, and RealWorldQA~\citep{xai2024realworldqa} — comparing against a broad set of pretrained vision-language models and assessing performance under two practical deployment scenarios: online routing and fusion without training data, and online routing and fusion with cross-domain training data available.

\textbf{Comparison with Pretrained Baselines.} 
Among the 11 pre-trained baselines, InternVL3-8B outperforms the other 10 VLMs on three of the four benchmarks with 57.76 on MMMU, 51.82 on MMMU-Pro and 88.47 on OKVQA. However, on RealWorldQA, InternVL3-8B (63.53) falls behine Qwen3-VL-8B (71.11) and InternVL2-8B (70.10). In comparison, 
Without access to any training data, VL-Scout (GRPO) outperforms all 11 VLMs in the pool on all four benchmarks, with 60.37 on MMMU, 52.01 on MMMU-Pro, 90.13 on OKVQA, and 76.47 on RealWorldQA, surpassing by a clear margin on MMMU and OKVQA. This result underscores a central thesis of our work: principled routing and fusion of pre-trained VLMs can be a strong alternative to scale individual model's capacity and yield consistent gains without any additional model training.

\vspace{-2pt}
\textbf{Routing-Only Strategies: The Case for REINFORCE+Prior.} For the family of routing-only (Top-1) methods, performance results vary considerably across bandit-based and policy gradient approaches, revealing important challenges 
in how well each strategy adapts to the online routing problem. Standard bandit methods, e.g., LinUCB, MA-TS, achieve moderate performance with 48.08 and 55.40 on MMMU, respectively, but struggle to capture the nuanced, query-dependent preferences that effective routing demands.
Policy gradient methods are 
more effective in this regard. VL-Scout (REINFORCE) improves upon the bandit baselines by a meaningful margin, while VL-Scout (REINFORCE+Prior) is the strongest routing-only method by a consistent margin across all 4 benchmarks, achieving 56.65 on MMMU, 50.129 on MMMU-Pro, 86.72 on OKVQA, and 68.15 on RealWorldQA. The incorporation of a prior over model preferences provides a crucial inductive bias that stabilizes early-stage routing decisions and accelerates convergence, particularly in low-data regimes where pure exploration is costly. These results establish REINFORCE+Prior 
as a 
competitive solution in settings where fusion is computationally prohibitive or architecturally constrained.

\vspace{-3pt}
\textbf{Routing + Fusion: VL-Scout (GRPO) Sets a New Standard.} 
The full potential of VL-Scout is realized through its fusion-based variants, which aggregate predictions across multiple models rather than committing to a single selection. Across all fusion strategies, VL-Scout (GRPO) achieves the state-of-the-art results in online without training data setting, with 60.37 on MMMU, 52.01 on MMMU-Pro, 90.13 on OKVQA, and 76.47 on RealWorldQA, outperforming all routing-only methods and all pretrained VLM baselines by substantial margins.
Compared to Voting-Top-3, VL-Scout (GRPO) improves MMMU by 2.73 points and RealWorldQA by 5.23 points, showing that naive ensemble voting is suboptimal and that learning to combine model outputs via group relative policy optimization yields meaningfully stronger aggregations. Compared to RLVR on RealWorldQA, the gains of GPRO are more pronounced (76.47 vs. 72.55), as GRPO's optimization objective better captures the distributional shift between benchmark domains.

\vspace{-3pt}
\textbf{Cross-Domain Training Further Amplifies VL-Scout's Advantages.\/} Incorporating cross-domain training data provides additional, consistent improvements across all VL-Scout variants, and VL-Scout (GRPO) leads all methods compared in Table~1 by a decisive margin, achieving 62.86 on MMMU, 53.52 on MMMU-Pro, 90.57 on OKVQA, and 74.51 on RealWorldQA. The improvement from the no-training setting is consistent across benchmarks, particularly notable on MMMU (+2.49 points) and MMMU-Pro (+1.51 points). It indicates that cross-domain routing signals transfer effectively via Prior and VL-Scout is capable of exploiting auxiliary supervision to refine its model selection strategy.
Also GRPO consistently outperforms RLVR, voting-based methods, REINFORCE+Prior, and bandit methods. 

\begin{figure*}[t]
    \centering
    \includegraphics[width=0.9\textwidth]{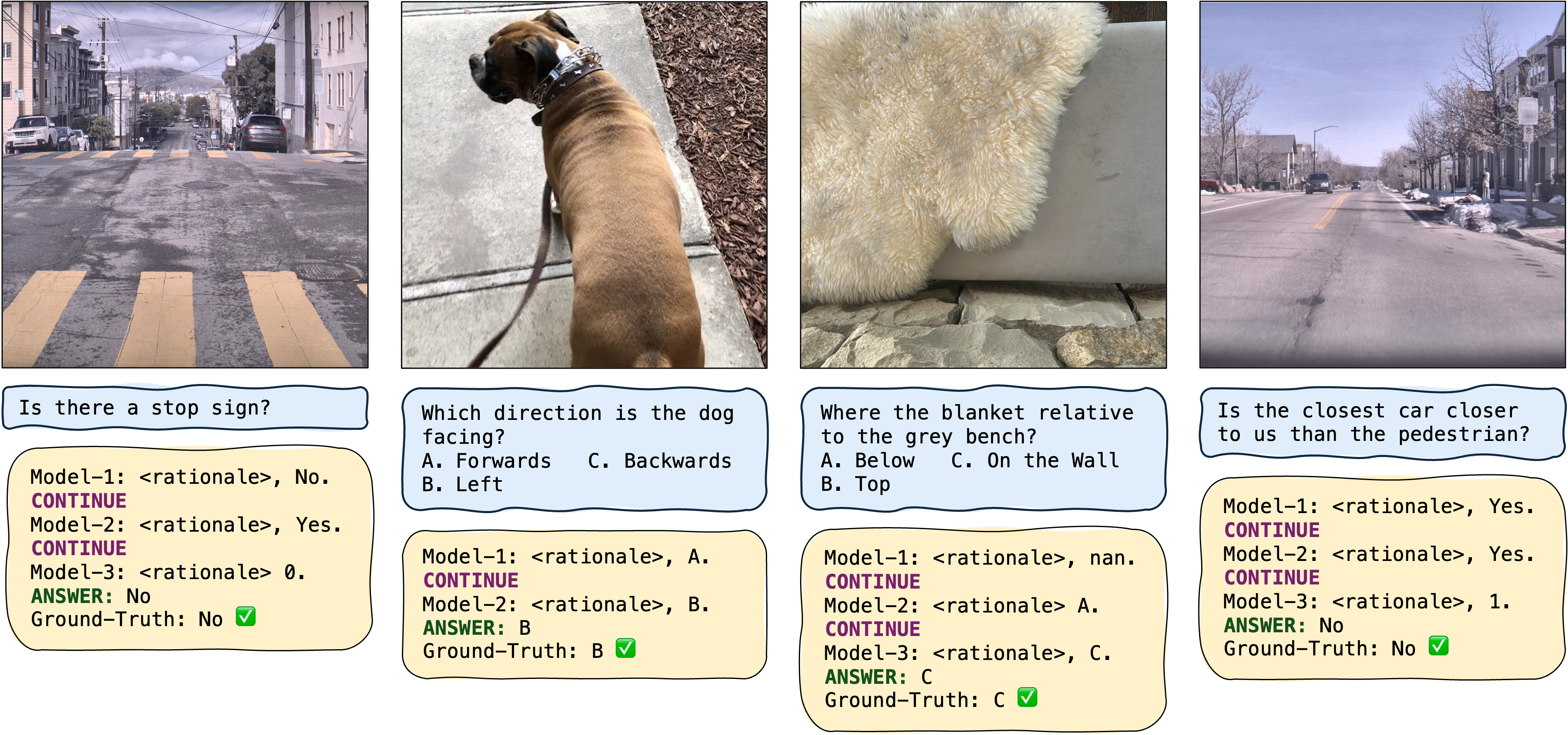}
    \caption{{Example illustration of VL-Scout reasoning. Example-1 reaches the correct answer after evaluating the first model's output. Example-2 makes the correct decision after consulting the second model. Example-3 continues consulting models until it finds a correct rationale and answer. Example-4 makes correct decision when all consulted models are incorrect. Detailed responses in Appendix-\ref{sec:examples}.}} 
    \label{fig:examples}
\end{figure*}

\textbf{Gap to Oracle.\/} Despite these strong results, a non-trivial gap remains between VL-Scout (GRPO) and the Oracle upper bound, which assumes perfect per-instance model selection and achieves 81.98 on MMMU, 78.14 on MMMU-Pro, 95.89 on OKVQA, and 92.54 on RealWorldQA. Even though there is a relatively high gap in MMMU and MMMU-Pro, VL-Scout achieves up to 5\% difference in OKVQA. Figure \ref{fig:examples} shows examples of how VL-Scout operates.
%
%

\begin{table}[t]
\centering
\vspace{-8pt}
\captionof{table}{VL-Scout against routing, collaboration, and ensemble baselines.}
\begin{adjustbox}{width=0.7\textwidth, center}
\begin{tabular}{l c c l c c}
    \toprule
    \multirow{2}{*}{Method} & \multirow{2}{*}{Routing} & \multirow{2}{*}{Fusion} & \multirow{2}{1.3cm}{Infer. Time (s)} & MMMU & A-OKVQA \\
    & & & & Acc (\%) & Acc (\%)\\
    \hline
    LLM-Blender (Pair R.)\citep{jiang2023llm} & \cmark & \xmark & 19.1s & 22.36 & 70.56 \\
    Majority Voting & \xmark & \cmark & 0.001 &  41.98 &  85.76 \\
    Plurality Voting & \xmark & \cmark & 0.001 &  52.54 &  86.81 \\
    RouteLLM (Random)\citep{ong2024routellm} & \cmark & \xmark & 0.0007 & 26.45 & 68.90 \\
    MAD w/ Consensus\citep{sivakumaran2025dart} & \xmark & \cmark & 21.84 & 53.2 & 57.4 \\
    MAD w/ Judge\citep{sivakumaran2025dart} & \xmark & \cmark & 27.92 & 54.8 & 59.7 \\
    DART (Best Model) \citep{sivakumaran2025dart} & \cmark & \cmark & 32.75 &  61.6 & 65.3 \\
    VL-Rethinker-7B \citep{wang2025vl} & \xmark & \xmark & 0.16 & 56.7 & 87.28 \\
    \hline
    VL-Scout & \cmark & \cmark & 4.09s & \textbf{62.86} & \textbf{90.13} \\
    \bottomrule
\end{tabular}
\end{adjustbox}
\label{tab:ensemble_methods}
\vspace{-6pt}
\end{table}

\vspace{-3pt}
\textbf{Performance against Prior Work.\/} 
Table~\ref{tab:ensemble_methods} benchmarks VL-Scout against a broad spectrum of multi-model inference strategies, and the results consistently favor VL-Scout across both accuracy and efficiency dimensions. Naive ensemble methods such as Majority Voting (41.98\%) and Plurality Voting (52.54\%) fall well short on MMMU despite their near-instantaneous inference times, demonstrating that undiscriminating aggregation fails to exploit the complementary strengths of heterogeneous VLMs. Multi-agent collaboration approaches close some of this gap, MAD w/ Judge and DART (Best Model) reach 54.8\% and 61.6\% on MMMU, respectively, yet with a prohibitive computational cost, requiring up to 32.75 seconds per query, nearly 8$\times$ the inference time of VL-Scout. Against VL-Rethinker-7B, VL-Scout delivers gains of 6.16 points on MMMU and 2.85 points on A-OKVQA. Ultimately, VL-Scout achieves state-of-the-art results on both benchmarks, 62.86\% on MMMU and 90.13\% on A-OKVQA, at just 4.09 seconds per query.

\begin{figure*}[hbt!]
    \centering
    \vspace{-6pt}
    \includegraphics[width=\textwidth]{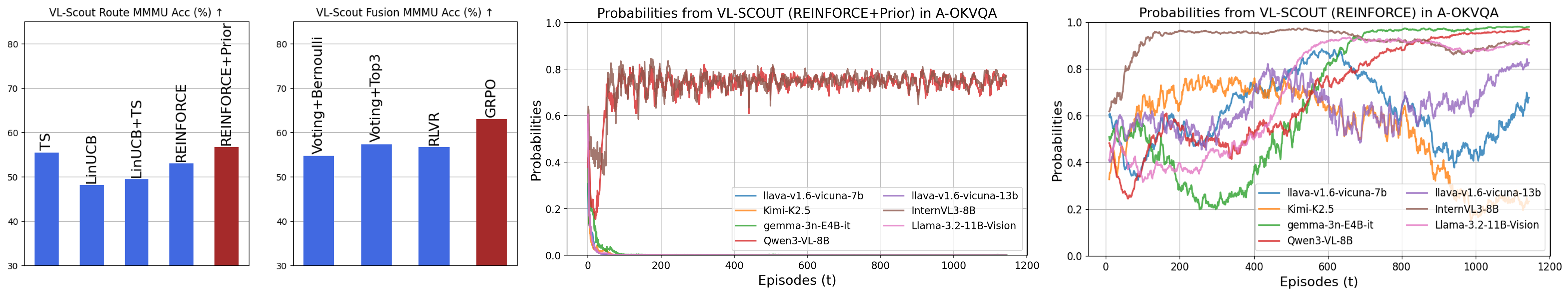}
    \vspace{-12pt}
    \caption{{The left shows an ablation study on routing and fusion in the MMMU dataset. The last two plots show how the prior affects the convergence on the stochastic model selection.}}
    \label{fig:ablation}
    \vspace{-8pt}
\end{figure*}

\begin{figure*}[hbt!]
    \centering
    
    \includegraphics[width=\textwidth]{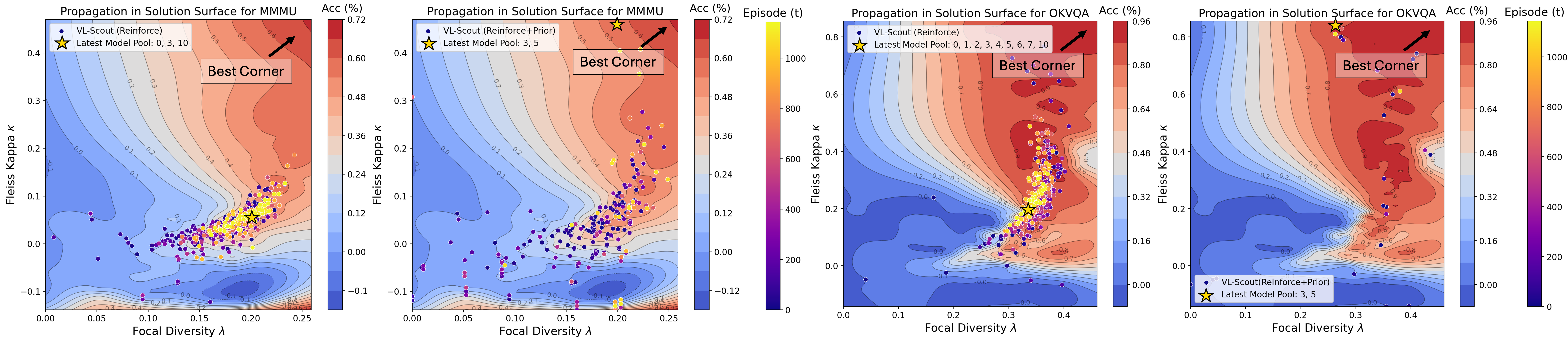}
    \caption{{All candidate ensemble teams from the model pool of size $K=11$ are plotted with their focal diversity scores\citep{tekin-etal-2024-llm}, Fleiss Kappa, and Plurality Voting Accuracy using two popular 
    datasets. We use cubic interpolation with RBF smoothing to create a surface, and the dark red represents a higher performance score. Each dot represents the explored ensemble team by the routing models REINFORCE and REINFORCE + PRIOR, colored by their appearances during testing.}}
    \label{fig:focal_first}
  \vspace{-10pt}
\end{figure*}

\vspace{-2pt}
{\bf Ablation Study.\/}
The bar chart in Figure \ref{fig:ablation} reports the results of experiments on the effect of Thompson sampling, linear context modeling, and non-linear context modeling of VL-Scout Route on the MMMU dataset. VL-Scout reaches its best performance when non-linear modeling of the context information is combined with Bayesian priors. This effect is also visible on the probability distribution change shown in the curve plots.
The prior regularizes the model selection distribution, whereas vanilla REINFORCE exhibits abrupt fluctuations. 
The second bar chart illustrates the impact of fusion modeling on performance, comparing Bernoulli sampling with voting, top-3 selection with voting, RLVR-based selection and fusion, and GRPO-trained fusion and selection.
Notably, VL-Scout Fusion with GRPO achieves the best performance. 
These results highlight the critical role of principled exploration and fusion strategies, demonstrating that combining non-linear context modeling with Bayesian priors and GRPO-based fusion yields the most stable and effective performance gains. Figure \ref{fig:focal_first} shows the effect of Prior to the REINFORCE, where the vanilla model cycles at the mediocre solution sets, Prior initialized REINFORCE algorithm quickly converges to the best model set selection. Without the prior, REINFORCE's sampled teams stay in a narrow band through the middle of the surface and end at mediocre teams on both benchmarks. With the Dirichlet prior, the router still explores widely in early episodes but then moves into the best corner and ends at the compact pair ${3,5}$ on both benchmarks; because these are the two models the operator prior up-weights, this reflects both the informative initialization and the KL regularization toward the belief.

\vspace{-8pt}
\section{Conclusion}
\vspace{-7pt}
 We presented VL-Scout, a unified stochastic routing framework for efficient multi-Modal and multi-step reasoning with turn-based reinforcement fusion. This paper makes three contributions. First, 
we combine Bayesian priors, policy gradient optimization, and sequential decision-making to enable structured exploration and adaptive inference. Second, a routing policy is trained via a KL-regularized REINFORCE objective by maximizing a reward-weighted evidence lower bound.
Third, we introduce a fusion policy trained with GRPO or RLVR, which selects between two actions: CONTINUE, invoking another model via the router, or ANSWER, terminating with a generated output. 
Evaluated on four representative multi-modal benchmarks (MMMU, MMMU-Pro, OKVQA, and RealWorldQA), compared against 11 representative routing and policy optimization baselines, we show that VL-Scout offers up to a 5\% 
improvement over the best individual model and prior-free routing baselines, while reducing premature policy collapse. 
We provide a reproducibility statement in Section \ref{sec:rep_statement},  more details on experimental setup and datasets in Appendix \ref{sec:setup}.


\subsection*{AI use statement}
In this work, we used generative AI tools for one task with required disclosure. We used Claude to help us to discover, navigate, and understand a proposition, for example, by explaining individual steps and clarifying the background results the proof relies on. We have not used generative AI tools for any other
task with required disclosure. Specifically, we have not used them to generate
synthetic datasets, develop theoretical models or conceptual frameworks,
propose or refine hypotheses, design or provide feedback on research
methodology or experiments, implement methods, clean or reformat datasets,
analyze data, or interpret results. Translation is not applicable to this work.

Additionally, we used generative AI tools (Claude and ChatGPT) for two tasks
with recommended disclosure. First, we used them to edit the paper to improve
readability. This included checking grammar, spelling, and phrasing in text
written by the authors, reformatting equations and tables, suggesting
corrections to mathematical notation, and drafting short connecting passages
(such as a section introduction and a transition sentence) that summarize
content already written by the authors. Second, we occasionally used them to
create or modify scientific figures, either by generating plotting code to
visualize results we had already obtained or by changing the visual style
(e.g., colors, fonts, layout) of existing figures. These tools did not generate
any of the research results or conclusions presented in this paper.

We have reviewed all AI-assisted work. Every suggested text edit was reviewed
and accepted or rejected by the authors to make sure the original meaning and
technical content were preserved, and every suggested notation correction was
checked against our derivations. We independently verified each step of the
proof and did not rely on AI-generated explanations without checking them. For
every AI-assisted figure, we checked the plotting code and verified that the
rendered figure accurately reflects the underlying experimental data. We take
responsibility for the final content of this work, including text, claims or
artifacts produced with the aid of generative AI.

\subsection{Reproducibility Statement}
\label{sec:rep_statement}

For VL-Scout implementation, a link to a downloadable source repository is included in our abstract. The source includes links for all the datasets, and we also provide the VLM outputs for each dataset. The details of our experiment are provided in Appendix \ref{sec:setup}, which includes the selected hyperparameters and hardware specifications. We also provide examples of the outputs and prompts used in our paper in Appendix \ref{sec:examples}.

\bibliography{iclr2027_conference}
\bibliographystyle{iclr2027_conference}

\newpage
{\bf {\Large Appendix}}

This appendix is organized by the following sections:

\section*{Appendix Contents}
\begin{itemize}
  \item Appendix \ref{sec:impact}: Impact Statement
  \item Appendix \ref{sec:design_guidelines}: Active Model Routing: Design Guidelines
  \item Appendix \ref{sec:proof}: Proof of Preposition 1
  \item Appendix \ref{sec:setup}: Dataset and Framework Parameters
  \item Appendix \ref{sec:classification}: Zero-shot Category Classification of Queries
  \item Appendix \ref{sec:prompts}: Prompts Used for VLMs and Fusion Model
  \item Appendix \ref{sec:examples}: Examples of VL-Scout Fusion

\end{itemize}

\appendix

\section{Impact Statement}
\label{sec:impact}
This paper presents work whose goal is to advance the field
of Machine Learning. There are many potential societal
consequences of our work, none which we feel must be
specifically highlighted here.

\section{Active Model Routing: Design Guidelines}
\label{sec:design_guidelines}

In this section, we build up the design of an active router step by step, starting from the simplest formulation and relaxing one assumption at a time. We consider a pool of $K$ vision-language models and a stream of queries $(x_t, \mathbf{I}_t)$ that arrive online, each consisting of a text prompt and an image. For every query, the router selects one or more models from the pool and receives a reward that may be delayed, partial, or noisy; no labeled training data or prior knowledge of the models is assumed. We first cast routing as a context-free Multi-Armed Bandit solved with Thompson Sampling, which learns each model's marginal accuracy but ignores the query itself. We then introduce query context through LinUCB, a contextual bandit with a linear reward model, and add stochastic exploration by combining it with Thompson Sampling. Finally, we replace the linear model with a neural policy trained via REINFORCE to capture non-linear relationships between queries and models. For each design, we discuss its limitations, which together motivate the design of VL-Scout Route.

\subsection{Multi-Armed Bandit (MA) and Thompson Sampling (TS).}
Based on our motivations in Section 2, we first approach the problem as a dynamic environment that can be modeled with only two elements: action and reward. From the user's perspective, we assume that there is no prior knowledge of the models in the pool or how they behave, and 
the queries are independent of each other. The problem is a natural fit to 
the Multi-Armed Bandit, where each arm represents a model we prompt. Let $a_t$ be the action at time $t$ and $r_t$ the reward (feedback), which may be delayed, partial, or noisy. The reward is given by $r_t = \mathcal{R}(a_t, y_t)+\epsilon_t$, where $\epsilon_t$ captures delay, partiality, or noise. For example, for binary feedback we define the reward function as
\begin{equation}
    \mathcal{R}(a_t, y_t) =
    \begin{cases}
        1 & \text{if } \hat{y}_t = y_t, \\
        0 & \text{otherwise},
    \end{cases}
    \label{eq:binary_reward}
\end{equation}
where $\hat{y}_t$ is the routed model's output.
In this setting, the goal is to maximize the cumulative reward over horizon $T$:
\begin{equation}
    \max \; \mathbb{E}\left[\sum_{t=1}^{T} r_t\right].
    \label{eq:cumulative_reward}
\end{equation}
This is equivalent to minimizing the regret, i.e., selecting the most rewarding model (arm) at each time step. As we do not know the models' reward distributions, Thompson Sampling~\citep{daniel2018tutorial} can be used to balance exploration and exploitation. At each query, we maintain a posterior over each arm's expected reward $\gamma_k$, modeled with a Beta distribution as the prior belief, $\gamma_k\sim \mathrm{Beta}(\alpha_k, \beta_k)$. The Beta distribution is also a natural fit when the reward is binary. For each arm $k\in\{1,\dots,K\}$ we maintain counts $\alpha_k$, the number of successes, and $\beta_k$, the number of failures of model $k$. For each query $x$, we sample a reward estimate from each arm's posterior and select the arm with the highest sample:
\begin{equation}
    k^{*} = \arg \max_{k} \gamma_k, \qquad \gamma_k \sim \mathrm{Beta}(\alpha_k, \beta_k).
    \label{eq:ts_select}
\end{equation}
After observing the reward $r\in\{0,1\}$, the posterior is updated by
\begin{equation}
    \alpha_k\leftarrow\alpha_k + r, \qquad \beta_k\leftarrow\beta_k + (1-r).
    \label{eq:ts_update}
\end{equation}
Therefore, we stochastically sample plausible models and act optimally with respect to the sample. After each observation, the posterior update makes the distribution wider or narrower, resulting in more exploration or exploitation, respectively. However, the caveat of this approach is that it is independent of the query context, which tends to fail under unseen queries and unknown context.  
Thompson Sampling~\citep{daniel2018tutorial} pools all the queries together, so $\alpha_k$ and $\beta_k$ converge to the model's marginal accuracy.

\subsection{Learnable Contextual Multi-Armed Bandits (LinUCB)}
The contextual information should drive model routing; for example, a vision model that excels at OCR questions may fail at spatial reasoning. Previous works \citep{nguyen2024metallm, ong2024routellm} showed that encoded semantic information of language queries can be used to decide which model should be queried. With a vision-language encoder, the input query can be mapped to a latent space, $g_{enc}(x, \mathbf{I})=\mathbf{h}$, $\mathbf{h}\in\mathbb{R}^d$. Based on this context, a bandit can learn the reward distribution to select models. Therefore, we employ a contextual bandit algorithm, LinUCB~\citep{li2010contextual}, which
selects actions by combining the predicted reward with an uncertainty bonus. It learns a linear function that maps context to expected reward for each model, and uses an upper confidence bound (UCB) to favor actions that are either promising or uncertain. This yields a principled exploration–exploitation trade-off with strong regret guarantees. For each action $k\in\{1,\dots, K\}$ and context $\mathbf{h}$, 
LinUCB assumes $\mathcal{R}(\mathbf{h};\theta_k)=\mathbf{h}^\top\theta_k$, where $\theta_k$ is the unknown coefficient vector. For each model $k$ we maintain the feature covariance matrix $\mathbf{A}_{k}\in\mathbb{R}^{d\times d}$ and the reward-weighted feature sum $\mathbf{b}_k\in\mathbb{R}^{d}$, and obtain the coefficients from the closed-form solution $\theta_{k}=\mathbf{A}_{k}^{-1}\mathbf{b}_k$. Given a new query $x_t$, the next arm is chosen with the UCB selection strategy:
\begin{equation}
    k^{*} = \arg \max_{k} \; \underbrace{\theta_{k}^{\top}\mathbf{h}}_{\text{predicted reward}} + \underbrace{\alpha \sqrt{\mathbf{h}^{\top}\mathbf{A}_{k}^{-1}\mathbf{h}}}_{\text{uncertainty}},
    \label{eq:linucb_select}
\end{equation}
where the uncertainty term is large when the context points in a direction the bandit has not yet explored. After observing reward $r_t$ for the selected arm, the update rule is:
\begin{equation}
    \mathbf{A}_k\leftarrow\mathbf{A}_k + \mathbf{h}\mathbf{h}^{\top}, \qquad \mathbf{b}_k\leftarrow\mathbf{b}_k+r_t\mathbf{h}, \qquad \theta_k\leftarrow\mathbf{A}_{k}^{-1}\mathbf{b}_k.
    \label{eq:linucb_update}
\end{equation}
LinUCB provides a strong baseline with theoretical regret guarantees under linear realizability assumptions. However, its assumption of a linear relationship between context and reward can be restrictive, particularly under cross-domain shifts. In addition, the UCB bonus is fully deterministic given the data, which may lead to the same arm being chosen whenever the same context is given, as there is no randomness at all. This means that the policy cannot explore when two arms have identical UCB scores. One of our design goals is to create a more robust algorithm against contextual domain shifts.

\subsection{LinUCB + Thompson Sampling}
We can replace the deterministic UCB score with a stochastic perturbation by sampling from a Bayesian posterior over the weight vectors:
\begin{equation}
    \tilde{\theta}_k\sim\mathcal{N}\!\left(\theta_k,\; v^{2}\mathbf{A}_k^{-1}\right),
    \label{eq:lints_sample}
\end{equation}
where $v^{2}\in \mathbb{R}$ scales the uncertainty. Each arm is then scored with its sampled weights, $\tilde{\theta}_{k}^{\top}\mathbf{h}$, and the update rule is the same as in Eq.~\eqref{eq:linucb_update}. As a result, we get randomized exploration from Thompson Sampling (TS) on top of the deterministic safety floor from UCB. 

\subsection{REINFORCE for Contextual Bandit Routing.}
There are two remaining concerns. First, the relationship between the context $\mathbf{h}$ and the action can be non-linear, which could be modeled via a multi-layer neural network with gradient-based optimization. Second, the previous algorithms use the immediate rewards in the update, whereas buffering rewards yields more stable updates. Therefore, we adopt REINFORCE \citep{williams1992simple}, a policy-gradient algorithm, for contextual bandit routing. Instead of maintaining a posterior over weights, it directly parameterizes a stochastic policy $\pi_{\theta}$ as a neural network and optimizes it via gradient ascent on the expected reward:
\begin{equation}
    J(\theta)=\mathbb{E}_{a\sim\pi_{\theta}(\cdot\mid x)}\left[\mathcal{R}(a,x)\right].
    \label{eq:reinforce_objective}
\end{equation}
Here, we use a two-layer MLP as the policy, with a sigmoid function $\sigma$ at the output layer:
\begin{equation}
    \pi_{\theta}(a_t\mid\mathbf{h}_t)= \sigma\!\left(\mathbf{W}_{L}\left(\dots \mathrm{ReLU}(\mathbf{W}_{1} \mathbf{h}_t)\dots\right)\right) = [p_1, p_2,\dots,p_K].
    \label{eq:reinforce_policy}
\end{equation}
The output has $K$ dimensions, where $p_k = \mathrm{Pr}(a_k=1\mid \mathbf{h}_t;\theta)$ is the probability of model $k$ being included in the selected pool, and action $a_k$ is drawn from a Bernoulli distribution with success probability $p_k$, i.e., $a_k\sim \mathrm{Bernoulli}(p_k)$. By the log-derivative trick, the policy gradient is estimated over a buffer of $T$ transitions:
\begin{equation}
    \nabla_{\theta} J(\theta) \approx \frac{1}{T}\sum_{t=1}^{T}\nabla_{\theta}\log\pi_{\theta}(a_t\mid\mathbf{h}_t)\, r_{t}.
    \label{eq:reinforce_grad}
\end{equation}
We also use reward normalization to reduce gradient variance. Compared to a Markov Decision Process, the contextual bandit has no transition model to learn, since each context is independent of its predecessor. 
However, REINFORCE leaves two issues open. First, the user who submits a query may have a prior belief about the models for some categories, e.g., `Deepseek-Math is more successful than Gemma-7B for math-science related tasks', which the policy cannot exploit. Second, due to the stochastic model selections made under noisy or delayed rewards, REINFORCE may converge prematurely to suboptimal models with insufficient exploration. 

\vspace{-3pt}
These design considerations serve as guidelines for developing our VL-Scout Route, 
which enables Thompson-style exploration while serving as a prior for regularization. Our experiments section will show the advantages of VL-Scout Route compared to the above four baselines.

\section{Proof of Proposition~\ref{prop:gibbs}}
\label{sec:proof}

\textbf{The normalizer $Z$.} The constant
\begin{equation*}
Z=\sum_{j=1}^{K}\tilde{p}_j\,e^{r(j)/\beta}=\mathbb{E}_{j\sim\tilde{\mathbf{p}}}\big[e^{r(j)/\beta}\big]
\end{equation*}
is the normalizing constant of $\pi^\star$: the prior-weighted average of the exponentiated rewards, which ensures $\sum_k\pi^\star(k)=1$. It depends on $\tilde{\mathbf{p}}$, $r$, and $\beta$, but not on $\pi$. Rearranging the definition of $\pi^\star$ gives, for every $k$,
\begin{equation}
\tilde{p}_k\,e^{r(k)/\beta}=Z\,\pi^\star(k).
\label{eq:gibbs_identity}
\end{equation}
For example, with two models, $\tilde{\mathbf{p}}=(0.5,0.5)$, $r=(1,-0.1)$, and $\beta=0.8$, we get $Z=0.5\,e^{1.25}+0.5\,e^{-0.125}\approx2.19$ and $\pi^\star\approx(0.80,0.20)$: the reward tilts the prior's even split toward the model that answered correctly.

\textbf{First statement.} Write both terms of the objective as sums over the models,
\begin{equation*}
\mathbb{E}_{\pi}[r]=\sum_k\pi(k)\,r(k),\qquad\beta\,\mathrm{KL}(\pi\|\tilde{\mathbf{p}})=\beta\sum_k\pi(k)\log\frac{\pi(k)}{\tilde{p}_k},
\end{equation*}
and write each reward as a logarithm, $r(k)=\beta\log e^{r(k)/\beta}$, so that $\mathbb{E}_{\pi}[r]=\beta\sum_k\pi(k)\log e^{r(k)/\beta}$. Both terms now have the form $\beta\sum_k\pi(k)\log(\cdot)$ and combine into a single sum:
\begin{equation*}
\mathbb{E}_{\pi}[r]-\beta\,\mathrm{KL}(\pi\|\tilde{\mathbf{p}})
=\beta\sum_k\pi(k)\Big[\log e^{r(k)/\beta}-\log\pi(k)+\log\tilde{p}_k\Big]
=-\beta\sum_k\pi(k)\log\frac{\pi(k)}{\tilde{p}_k\,e^{r(k)/\beta}} .
\end{equation*}
The expected reward has not disappeared: it is carried by the factor $e^{r(k)/\beta}$ in the denominator. Substituting Eq.~(\ref{eq:gibbs_identity}) and splitting the logarithm gives
\begin{equation*}
-\beta\sum_k\pi(k)\log\frac{\pi(k)}{Z\,\pi^\star(k)}
=-\beta\sum_k\pi(k)\log\frac{\pi(k)}{\pi^\star(k)}+\beta\log Z\sum_k\pi(k)
=\beta\log Z-\beta\,\mathrm{KL}(\pi\|\pi^\star),
\end{equation*}
where the last step uses $\sum_k\pi(k)=1$. Since $Z$ does not depend on $\pi$, and $\mathrm{KL}(\pi\|\pi^\star)\ge 0$ with equality if and only if $\pi=\pi^\star$, $\pi^\star$ is the unique maximizer. The maximum value, $\beta\log Z$, is a soft maximum of the rewards under the prior: it lies between $\mathbb{E}_{k\sim\tilde{\mathbf{p}}}[r(k)]$ (by Jensen's inequality) and $\max_k r(k)$.

\textbf{Second statement.} Let $P(O{=}1\mid k)=e^{r(k)/\beta}/C$ with $C\ge\max_k e^{r(k)/\beta}$, so that these are valid probabilities. Then $P(O{=}1)=\sum_k\tilde{p}_k\,P(O{=}1\mid k)=Z/C$, so $\log P(O{=}1)=\log Z-\log C$, and for every $\pi$, by Jensen's inequality,
\begin{equation*}
\log P(O{=}1)\ \ge\ \mathbb{E}_{\pi}\big[\log P(O{=}1\mid k)\big]-\mathrm{KL}(\pi\|\tilde{\mathbf{p}})=\frac{1}{\beta}\Big(\mathbb{E}_{\pi}[r]-\beta\,\mathrm{KL}(\pi\|\tilde{\mathbf{p}})\Big)-\log C,
\end{equation*}
with equality if and only if $\pi(k)=P(k\mid O{=}1)=\tilde{p}_k\,e^{r(k)/\beta}/Z=\pi^\star(k)$. \hfill$\square$

\section{Dataset and Framework Parameters}
\label{sec:setup}
{\bf Datasets.\/} The experiments are performed on four popular benchmark datasets: MMMU\cite{yue2024mmmu}, MMMU-Pro \cite{yue2024mmmu_pro}, A-OKVQA \cite{schwenk2022okvqa}, and RealWorldQA~\cite{xai2024realworldqa}.
in multiple-choice question format. MMMU\cite{yue2024mmmu} dataset is designed to evaluate VLMs on massive multi-discipline tasks containing 30 different subjects. The questions are at the college level and contain multiple reasoning steps. The dataset only provides 5 development and 900 validation samples with the labels. However, the labels of the test split are hidden. MMMU-Pro dataset~ \cite{yue2024mmmu_pro}, on the other hand, is the robust version of MMMU and provides only a test dataset with labels. A-OKVQA dataset~\cite{schwenk2022okvqa} provides visuals with questions that require reasoning to solve. The expected answer is either the choice from the multiple choices given or a direct answer. There are 17,056 training pairs, 1,146 validation pairs, and 6,70 test samples.  We use the validation set for evaluation. Similar to the MMMU dataset, the A-OKVQA benchmark did not share the test labels. Thus, we used a validation split as the test data. RealWorldQA is a benchmark designed for real-world understanding. The dataset consists of anonymized images taken from vehicles, in addition to other real-world images. It contains spatial understanding that requires OCR and object detection. It contains 764 images and questions, where questions have 3-4 choices.

\textbf{Evaluation Metrics.} To evaluate the performance of VLMs and V3Fusion methods, we use accuracy on MMMU, MMMU-Pro and A-OKVQA datasets.

\textbf{VLM selection for the candidate model pool.} In our model pool selection, we choose pre-trained individual VLMs based on the following three elements and their effect on performance: (i) size of the model (less than 16b), (ii) model variety, and (iii) being open-source. The size was the limiting factor for us since the modern LLMs are huge, and our experiments were conducted on a single H100 GPU of 80GB.

\textbf{VL-Scout Policy Models and Structural Details.} 
VL-Route REINFORCE model contains two fully connected hidden layers with 100 neurons and ReLU activations between the layers. The model weights start from Xavier initialization and are optimized by Adam during online training. For the VL-Scout Fusion, we have used \texttt{Qwen/Qwen3-VL-4B-Instruct} and during training we have used LORA with rank $8$ and $\mathrm{lora}\_\alpha=16$ and trained with Decoupled Adam.

\begin{table}[h]
\centering
\caption{Question Categories and Descriptions}
\label{tab:categories}
\begin{tabular}{ll}
\toprule
\textbf{Category} & \textbf{Description} \\
\midrule
\texttt{spatial\_and\_visual}      & Where, position, layout, scene understanding \\
\texttt{counting\_and\_quantity}   & How many, amounts, numbers \\
\texttt{text\_and\_document}       & OCR, reading, documents, charts \\
\texttt{science\_and\_math}        & Physics, chemistry, biology, math, engineering \\
\texttt{humanities\_and\_arts}     & History, literature, music, art, philosophy \\
\texttt{social\_and\_applied}      & Economics, law, medicine, psychology, business \\
\texttt{perception\_and\_attribute}& Color, shape, material, appearance, action \\
\texttt{knowledge\_and\_reasoning} & World knowledge, causal, comparison, why \\
\bottomrule
\end{tabular}
\end{table}

\section{Zero-shot Category Classification of Queries}
\label{sec:classification}

The Table \ref{tab:categories} shows the main categories and the descriptions that fall under each category. We have used \texttt{nli-MiniLM2-L6-H768} zero-shot language classifier to classify each question into one of the categories.

\begin{table}[h]
\centering
\caption{Prompt Configurations}
\label{tab:prompts}
\begin{tabular}{p{2.5cm} p{4cm} p{7cm}}
\toprule
\textbf{Prompt} & \textbf{Description} & \textbf{Content} \\
\midrule

\texttt{VQA System Prompt}
&
Base prompt. Instructs the model to output the final answer only, no reasoning.
&
\texttt{You are a visual question answering assistant.} \\
& & \texttt{For every question, first write a short rationale } \\
& & \texttt{describing what you observe and how you reason, then give } \\
& & \texttt{your final answer. Always use this format:} \\
& & \texttt{Rationale: [your reasoning]} \\
& & \texttt{Answer: [your final answer]]} \\

\midrule

\texttt{Decision Agent Prompt}
&
Dynamic multi-model prompt. Decides whether to STOP and answer or CONTINUE querying more models.
&
\texttt{You are a decision agent that determines whether} \\
& & \texttt{to STOP and answer, or CONTINUE querying more models.} \\
& & \texttt{Query: \{query\}} \\
& & \texttt{Responses collected so far:} \\
& & \quad \texttt{Model \{i+1\}: Answer: \{resp[`answer']\}} \\
& & \quad \texttt{Rationale: \{resp[`rationale']\}} \\
& & \texttt{Models remaining in pool: \{remaining\}} \\
& & \texttt{Think about whether the current responses are sufficient:} \\
& & \quad \texttt{- Do the models agree?} \\
& & \quad \texttt{- Is the rationale strong and complete?} \\
& & \quad \texttt{- Would another model's perspective help?} \\
& & \texttt{If sufficient $\rightarrow$ ANSWER: <final answer>} \\
& & \texttt{If not $\rightarrow$ CONTINUE} \\

\bottomrule
\end{tabular}
\end{table}

\section{Prompts Used for VLMs and Fusion Model}
\label{sec:prompts}

The Table \ref{tab:prompts} shows the prompts we have used for inferring the base models in the pool (first row) and the prompt for the VL-Scout Fusion model to decide on CONTINUE or ANSWER (second row).  

\begin{table}[h]
\centering
\caption{Model Rationales and Answers for Example-4 for Figure-\ref{fig:examples}}
\label{tab:model_responses}
\begin{tabular}{p{2.5cm} p{7.5cm} p{1.2cm} p{1.5cm} p{1.5cm}}
\toprule
\multicolumn{5}{l}{\textbf{Question:} Is the closest car closer to us than the pedestrian?} \\
\midrule
\textbf{Model} & \textbf{Rationale} & \textbf{Answer} & \textbf{Decision} & \textbf{Correct?} \\
\midrule

\texttt{Qwen} (Model 1)
&
The pedestrian is visible on the sidewalk to the right, standing near a tree and some snow piles. The closest car is further down the street, behind the pedestrian. Since the pedestrian is on the sidewalk and the car is on the road and further away, the pedestrian is closer to the viewer than the car.
&
Yes
&
CONTINUE
&
\textcolor{wrongred}{$\times$} \\

\midrule

\texttt{InternVL} (Model 2)
&
The closest car is on the left side of the road, while the pedestrian is on the right side. By comparing their relative positions to the camera's viewpoint, the car appears closer to the viewer than the pedestrian.
&
Yes
&
CONTINUE
&
\textcolor{wrongred}{$\times$} \\

\midrule

\texttt{Gemma} (Model 3)
&
The image shows a street scene. The closest car is driving down the middle of the road, relatively far away. The pedestrian is walking on the sidewalk on the right side, and is much closer to the camera than the car. Therefore, the pedestrian is closer to us than the closest car.
&
1
&
ANSWER: No
&
\textcolor{correctgreen}{\textbf{\checkmark}} \\

\midrule

\multicolumn{2}{l}{\textbf{Ground Truth}}
&
\multicolumn{2}{l}{\textbf{No}}
&
\textcolor{correctgreen}{\textbf{\checkmark}} \\

\bottomrule
\end{tabular}
\end{table}

\begin{table}[h]
\centering
\caption{Model Rationales and Answers Example-1 for Figure-\ref{fig:examples}}
\label{tab:model_responses_stopsign}
\begin{tabular}{p{1.8cm} p{7.5cm} p{1.2cm} p{1.5cm} p{1.5cm}}
\toprule
\multicolumn{5}{l}{\textbf{Question:} Is there a stop sign?} \\
\midrule
\textbf{Model} & \textbf{Rationale} & \textbf{Answer} & \textbf{Decision} & \textbf{Correct?} \\
\midrule

\texttt{InternVL} (Model 1)
&
To determine if there is a stop sign, I examine the street for any octagonal signs with a red border and white background. Upon reviewing the image, I see several signs along the street, but none match the description of a stop sign. The signs visible are either rectangular or do not have the characteristic red border and white center.
&
No
&
CONTINUE
&
\textcolor{correctgreen}{\textbf{\checkmark}} \\

\midrule

\texttt{Qwen} (Model 2)
&
I can see a white SUV parked on the left side of the street, and behind it, there appears to be a stop sign mounted on a pole near the sidewalk. The sign is partially obscured by the SUV and the angle of the photo, but its shape and position are consistent with a stop sign. Therefore, based on the visual evidence, there is at least one stop sign present.
&
Yes
&
CONTINUE
&
\textcolor{wrongred}{$\times$} \\

\midrule

\texttt{Gemma} (Model 3)
&
I see a street with cars parked on the side and yellow crosswalk markings. I do not see any stop signs visible in the image.
&
0
&
ANSWER: No
&
\textcolor{correctgreen}{\textbf{\checkmark}} \\

\midrule

\multicolumn{2}{l}{\textbf{Ground Truth}}
&
\multicolumn{2}{l}{\textbf{No}}
&
\textcolor{correctgreen}{\textbf{\checkmark}} \\

\bottomrule
\end{tabular}
\end{table}

\section{Examples of VL-Scout Fusion}
\label{sec:examples}

We show the rationales and answers by each base VLM in Tables \ref{tab:model_responses_stopsign} and \ref{tab:model_responses} for cases in Figure \ref{fig:examples}. These two examples illustrate how the sequential querying strategy handles both model disagreement and majority error. In the first example (closest car vs. pedestrian), Models 1 and 2 confidently agree on the wrong answer, yet Model 3 correctly identifies the depth relationship, steering the decision agent toward the correct final answer. In the second example (stop sign detection), Model 2 hallucinates a partially obscured stop sign, while Models 1 and 3 independently and correctly find none present. In both cases, the decision agent successfully resolves conflicts and overrides incorrect responses by continuing to query additional models, demonstrating the system's robustness against individual model failures and hallucinations.


\end{document}